\pdfoutput=1

\documentclass[11pt]{article}

\usepackage[]{acl}

\usepackage{times}
\usepackage{latexsym}
\usepackage{graphicx}
\usepackage{subcaption}
\usepackage[T1]{fontenc}

\usepackage[utf8]{inputenc}

\usepackage{microtype}
\usepackage{adjustbox, booktabs, multirow, pbox}
\usepackage{color, colortbl, xcolor}
\usepackage{amsmath}
\usepackage{amssymb}
\usepackage{amsthm}
\usepackage{amsfonts}
\usepackage{bm}

\newcommand*{\Rb}{\mathbb{R}}

\newcommand*{\matr}[1]{\mathbf{#1}}
\newcommand*{\vect}[1]{\bm{#1}}

\newcommand{\ignore}[1]{}

\title{Structured Prompt Tuning}

\author{
  Chi-Liang Liu$^\dagger$ \quad Hung-yi Lee$^\dagger$ \quad Wen-tau Yih$^\ddagger$ \\[0.5ex]
  $^\dagger$National Taiwan University \quad $^\ddagger$Meta AI\\
  \texttt{\{r07942083, hungyilee\}@ntu.edu.tw} \quad
  \texttt{scottyih@fb.com} \\
  }

\begin{document}
\maketitle

\begin{abstract}

We propose \emph{structured prompt tuning}, a simple and effective method to improve prompt tuning. 
Instead of prepending a sequence of tunable embeddings to the input, we generate the soft prompt embeddings through a hypernetwork.
Our approach subsumes the standard prompt tuning, allows more flexibility in model design and can be applied to both single-task and multi-task training settings.
Empirically, structured prompt tuning shows a gain of +1.2$\sim$1.5 points on the GLUE benchmark and is less sensitive to the change of learning rate, compared to standard prompt tuning.



\end{abstract}

\section{Introduction}
Prompting large language models has been shown an effective way to avoid expensive full model fine-tuning on a wide range of downstream NLP tasks.
This approach prepends either natural language texts or continuous embeddings to the input. 
For example, \citet{gpt3} demonstrated that pretrained language models are able to do in-context learning, where the model adapts to a new task simply by prepending a few training examples. 
In contrast to prompting with textual tokens, soft prompt tuning~\citep{qin-eisner-2021-learning, li2021prefixtuning, lester2021power}, which prepends a few tunable embeddings to the inputs, has also been proposed.

Despite its effectiveness, soft prompt tuning is fundamentally limited in its form of interacting with the language models. 
Variants of this approach usually differ only in where the soft prompt tokens are placed.
For example, \citet{lester2021power} put the soft prompt tokens before the input; \citet{hambardzumyan-etal-2021-warp} insert prompt tokens before, between and after the sentences; \citet{zhong-etal-2021-factual} concatenate soft prompt tokens after the sentence.
Adding soft prompt tokens in different Transformer layers has also been attempted~\cite{qin-eisner-2021-learning,li2021prefixtuning}.

\begin{figure}[t]
    \centering
    \includegraphics[width=\columnwidth]{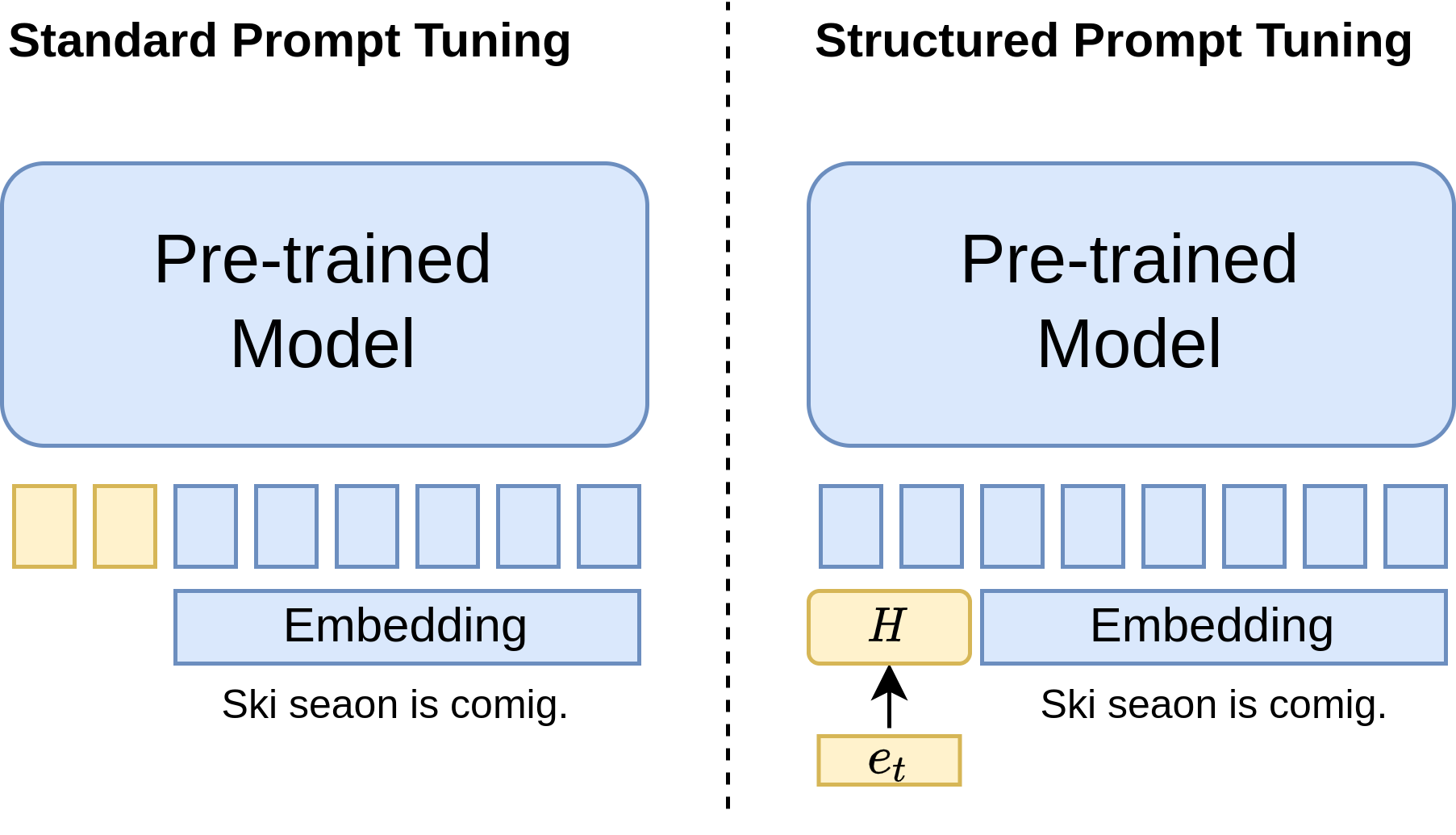}
    \caption{\textbf{Standard prompt tuning} stores soft prompts for each tasks. \textbf{Structured prompt tuning} generates soft prompts with the task embedding $e_{t}$ and hypernetwork network $H$. Yellow boxes indicate trainable parameters; blue boxes indicate frozen parameters.}
    \label{fig:cont}
\end{figure}

In this paper, we propose \emph{structured prompt tuning}, where the soft prompts are \emph{generated} by a hypernetwork that takes as input a \emph{task} embedding, as illustrated in Figure~\ref{fig:cont}.
Structured prompt tuning generalizes the standard soft prompt tuning method in that with a particular hypernetwork architecture, structured prompt tuning is in fact equivalent to soft prompt tuning. 
Perhaps more importantly, structured prompt tuning provides additional flexibility in model design, as different hypernetwork architectures impose implicit structures among soft prompt embeddings.
Such flexibility can be crucial in better adapting the model to the target task.

With simple hypernetwork architectures, such as linear layer, low rank linear layer or a multilayer perceptron, our structured prompt tuning consistently outperforms standard prompt tuning on various NLU tasks, such as question answering and sentiment classification in the single-task training setting. 
For the multi-task training setting, our proposed method surpasses not only standard prompt tuning but also the full fine-tuning approach. 
We additionally find that structured prompt tuning is less sensitive to learning rate change, compared to the standard prompt tuning.


\ignore{

This allows us with more ways to interact with the model since the design of the hypernetwork does not have any limitations as long as it can generate the soft prompt we requires. 
Standard soft prompt tuning can be considered as a special case of structured prompt tuning. 

We explore a few simple architectures for the hypernetwork, including linear layer, low rank linear layer, and a mulitlayer perceptron in our work.

Figure~\ref{fig:cont} provides a difference between standard prompt tuning and structured prompt tuning.

Our structured prompt tuning consistently outperforms standard prompt tuning over various NLU tasks such as question answering and sentiment classification with a single-task training setting. For the multi-task training setting, our proposed method surpasses not only standard prompt tuning but also fully finetune's performance. We additionally find that structured prompt tuning is less sensitive to learning rate compared to the standard prompt tuning. The qualitative study indicates that the improvement is not due to the initialization of parameters.

}

\section{Related Work} 

Transfer learning by adapting large pre-trained language models~\citep{devlin-etal-2019-bert, liu2019roberta, lewis-etal-2020-bart, 2020t5} to specific tasks has been the standard approach in NLP.
While full fine-tuning is the common adaptation method, it requires storing and updating all model parameters. 
As a result, more efficient alternatives have been proposed by researchers.

\citet{gpt3} demonstrated that PLMs can achieve surprising results on downstream tasks by prepending a few training examples to the model input without tuning the model parameters. 
This method has been referred to as prompting or in-context learning.
However, the performance is sensitive to its prompts~\citep{gpt3,schick-schutze-2021-exploiting,schick-schutze-2021-just,shin-etal-2020-autoprompt}, which triggers research on designing or learning appropriate prompts.
For instance, AutoPrompt~\citep{shin-etal-2020-autoprompt} leverages gradient-guided search over the large discrete space of phrases to construct a prompt. 
\citet{gao2021making} use T5~\citep{2020t5} to generate template candidates to obtain high-quality prompts guided by the development set performance.
Instead of optimizing in the discrete tokens space, ``soft prompt'' methods that replace discrete tokens with randomly initialized continuous embeddings and update them by gradient descent have been studied. 
For example, \citet{zhong-etal-2021-factual} and \citet{qin-eisner-2021-learning} explore the possibility of using cloze prompts
to query pre-trained language models for single-word answers. 
\citet{li2021prefixtuning} and \citet{lester2021power} extend the idea to generation tasks and show that soft-prompt tuning is competitive to full model fine-tuning. 
Concurrent to our work, \citet{standing} introduce input-dependent prompt tuning that also generates prompt tokens using a generator. 
In contrast to their approach, our generator takes as input a task embedding instead of input tokens. 

\section{Structured Prompt Tuning}

\begin{table*}[t]
    \centering
    \setlength\tabcolsep{5pt}
    \resizebox{\textwidth}{!}{%
    \begin{tabular}{lllllllllll}
     \toprule
     & \multicolumn{2}{c}{\bf{Single Sentence}} & \multicolumn{3}{c}{\bf{Sentence Similarity}} & \multicolumn{3}{c}{\bf{Natural Language Inference}} & &\bf{QA}\\
     & \bf{SST-2} & \bf{COLA} & \bf{MRPC} & \bf{STS-B} & \bf{QQP} & \bf{MNLI} & \bf{QNLI} & \bf{RTE} & \bf{Avg.} &  \bf{SQuAD}\\
     \midrule
      \textit{Training Size} & \it66k & \it8.5k & \it3.7k & \it7k & \it363k & \it392k & \it107k & \it2.5k & - & 87k\\
     \midrule
     \multicolumn{11}{l}{\it \textbf{T5-Small}}\\
     Prompt Tuning  &   $88.9_{0.4}$ & $40_{2.6}$ & $84.2/89.2_{0.5/0.2}$ & $\bf86.2/86.1_{0.8/1.0}$ & $79.8/75.1_{0.5/0.4}$ & $67.2_{0.4}$ & $82.4_{0.5}$ & $59.7_{5.1}$ & $73.5_{1.2}$ & $49.0/60.5_{0.5/0.2}$ \\
 SPT - Linear  &  $89.0_{0.6}$ & $43.8_{1.7}$ & $85.2/89.6_{1/0.8}$ & $85.9/85.6_{0.3/0.2}$ & $\bf81.6/76.6_{0.3/0.1}$ & $67.4_{0.1}$ & $83.1_{0.3}$ & $\bf61.6_{5.2}$ & $74.6_{0.8}$ & $\bf 49.9/61.2_{0.4/0.4}$ \\
 SPT - LowRank  & $88.5_{0.9}$ & $41.2_{2.9}$ & $\bf86.5/90.5_{1.4/0.7}$ & $85.6/85.4_{0.9/0.8}$ & $81.9/75.9_{0.3/1.0}$ & $\bf67.7_{0.2}$ & $83.1_{0.5}$ & $60.4_{5.2}$ & $74.2_{1.0}$ & $49.5/60.9_{0.1/0.2}$ \\
 SPT - MLP  & $\bf89.4_{0.3}$ & $\bf44.1_{5.3}$ & $86.4/90.3_{1.2/1.0}$ & $84.8/84.8_{0.4/0.3}$ & $\bf81.9/76.3_{0.5/0.5}$ & $67.4_{0.4}$ & $\bf83.5_{0.7}$ & $63.3_{2.9}$ & $\bf75.0_{0.3}$ & $49.0/60.3_{0.9/0.3}$ \\
 Fully Finetune & $91.3_{0.5}$ & $48.3_{3.4}$ & $84.9/89.4_{1.4/0.9}$ & $83.7/83.3_{0.7/0.7}$ & $90.4/87_{0.1/0.1}$ & $81.8_{0.7}$ & $88_{1.8}$ & $58.2_{1.7}$ & $78.4_{0.3}$ & $72.1/81.8_{0.3/0.1}$\\
     \midrule
     \multicolumn{11}{l}{\it \textbf{T5-Base}}\\
 Prompt Tuning  &     $92.7_{0.4}$ & $53.9_{2.2}$ & $85.4/89.9_{2.1/1.2}$ & $89.2/89.2_{0.7/0.8}$ & $83.2/78.2_{0.2/0.6}$ & $77.9_{0.8}$ & $88.7_{0.9}$ & $\bf63.0_{1.3}$ & $79.2_{0.5}$ & $68.9/78.7_{2.1/2}$\\
 SPT - Linear  &      $92.9_{0.5}$ & $54.8_{2.4}$ & $87.8/91.5_{1/0.5}$ & $87.9/88_{0.7/0.5}$ & $84.1/79.8_{0.6/0.8}$ & $80.3_{0.3}$ & $88.7_{0.6}$ & $60.1_{1.3}$ & $79.6_{0.4}$ & $70.9/80.6_{0.4/0.1}$\\
 SPT - LowRank  &     $\bf93.1_{0.2}$ & $56.2_{1.7}$ & $86/90.2_{2.5/1.9}$ & $\bf89.5/89.3_{0.7/0.7}$ & $\bf85.2/80.7_{0.3/0}$ & $\bf80.7_{0.3}$ & $\bf89.9_{0.3}$ & $62.8_{2.5}$ & $\bf80.4_{0.5}$ & $70.8/80.4_{0.3/0.2}$\\
 SPT - MLP  &  $93.1_{0.4}$ & $\bf57.5_{1.0}$ & $\bf88.3/91.8_{1.1/0.8}$ & $88.9/88.9_{0.9/0.8}$ & $\bf85.1/80.8_{0.4/0.4}$ & $80.5_{0.5}$ & $89.1_{0.3}$ & $56.5_{3.8}$ & $79.8_{0.4}$ & $\bf71.0/80.6_{0.5/0.6}$\\
     Fully Finetune  & $93.8_{0.1}$ & $56.2_{3.4}$ & $90.3/92.3_{2/0.3}$ & $89.7/89.5_{0.5/0.4}$ & $91/87.9_{0.1/0.1}$ & $87.3_{0.2}$ & $91.7_{0.4}$ & $66.9_{0.8}$ & $83.3_{0.4}$ & $74.2/83.9_{0.3/0.3}$\\
    
    %
    \bottomrule
    \end{tabular}}
    \caption{\textbf{GLUE and SQuAD test set results.}  SPT stands for structured prompt tuning. As in the original GLUE paper,  we report the Matthews correlation coefficient for CoLA. We report accuracy and F1 for MRPC and QQP. Pearson and Spearman correlations are provided for STS-B. MNLI accuracy is reported on matched test sets.  For all other tasks, we report accuracy. The \textit{Avg.} column shows the GLUE scores averaged across all tasks with equal weights. Numbers in the subscript indicate the standard deviation across 3 random seeds. \textbf{Bold} indicates the best results among the prompt tuning methods.} 
    \label{tab:main_results}
\end{table*}

As illustrated in Figure~\ref{fig:cont}, our proposed \textit{structured} prompt tuning approach is similar to soft prompt in that the origin input is prepended with $n$ token embeddings.
However, these embeddings are not trained directly, but are instead generated by a hypernetwork that takes a \emph{task} embedding as input.
The tuning procedure updates the hypernetwork parameters and the task embedding.
We describe our method formally in this section.


\subsection{Method}
Following T5~\citep{2020t5} and prompt tuning~\citep{lester2021power}, we reduce any downstream task to a seq2seq text generation task. 
Given a series of input tokens $\matr{X}$ and a sequence of tokens $\matr{Y}$ that represents a class label for classification tasks or a ground truth sequence for generation tasks, our goal is to maximize the the conditional probability $P_{\textrm{LM}}(\matr{Y}|\matr{X})$, where $P_{\textrm{LM}}$ is a language model.

\paragraph{Prompt Tuning} Following \citet{lester2021power}, we prepend $n$ prompt representations $\vect{r}_1\dots \vect{r}_n$ to the input, where $\vect{r}_i \in \Rb^d$ and $d$ is dimension of the embeddings. 
The new conditional probability is thus calculated by $P_{\textrm{LM}}(\matr{Y}|\vect{r}_1\dots \vect{r}_n,\matr{X})$.
Notice that unlike full model fine-tuning, the parameters in the original model, $P_{\textrm{LM}}$, are frozen. 
Only the prompt representations $\vect{r}_1 \dots \vect{r}_n$ are tunable.


\paragraph{Structured Prompt Tuning} 
Standard prompt tuning learns the prompt representations $\vect{r}_1\dots \vect{r}_n$ directly.
In contrast, we \emph{generate} these representations via a hypernetwork $H$:
\begin{equation}
    \matr{R} = H(\vect{e_{t}}), 
\end{equation}
where $\matr{R} = [\vect{r}_1;\vect{r}_2;\dots;\vect{r}_n]\in \Rb^{n\times d}$ is the prompt representations and $\vect{e}_{t} \in \Rb^{k}$ is the task embedding for task $t$. 
The hypernetwork $H$ and the task embedding $\vect{e}_{t}$ are the only trainable parameters.
The design of task embedding $\vect{e}_{t}$ makes our prompt tuning method applicable to both single- and multi-task learning scenarios.
When there are more than one task to be considered, each task is represented by a different embedding $\vect{e}_t$, but $H$ is shared by all the tasks. Note that standard prompt tuning is a special case of structure prompt tuning with a $1 \times nd$ matrix as the hypernetwork, and $k = 1$ for the task embedding dimension.


\subsection{Hypernetwork Architecture}
We consider three different architectures for the hypernetwork. 
To simplify the description below, we introduce a matricization\footnote{Equivalent to \texttt{reshape(vector, (batch\_size, n\_tokens, dimension))} in numpy.} notation $\mathcal{M}: \Rb^{nd}\rightarrow \Rb^{n\times d}$, which constructs a matrix from a vector.

\paragraph{Linear}
We linearly project the shared vector to different representation
subspaces. 
\begin{equation}
    H(\vect{e}_{t})  = \mathcal{M}({\matr W\vect{e}_{t} + \vect b})
\end{equation}
where $\matr W \in \Rb^{nd \times k}$ and $\vect b \in \Rb^{nd}$.

\paragraph{Low-Rank Linear}
To strengthen the relationship between generated prompts, we introduce a low-rank constraint to the hypernetwork: matrix $\matr{W}$ is derived via low-rank factorization. 
\begin{equation}
    H(\vect{e}_{t})  = \mathcal{M}({\matr{W}\vect{e}_{t} + \vect b}) = \mathcal{M}({\matr{C}\matr{F}\vect{e}_{t} + \vect b}),
\end{equation}
where $\matr C \in \Rb^{nd \times r}, \matr F \in \Rb^{r \times k}$ and $r$ is a hyperparameter that specifies the rank of $\matr W$.

\paragraph{Multilayer Perceptron (MLP)} 
We also try an MLP model that consists of two linear transformations with a GeLU activation $\phi$ in between.
\begin{equation}
    H(\vect{e}_{t})  = \mathcal{M}({\matr{W}_2 \phi (\matr{W}_1\vect{e}_{t} + \vect{b}_1}) + \vect{b}_2),
\end{equation}
where $\matr{W}_1 \in \Rb^{h \times k}, \matr{W}_2 \in \Rb^{nd \times h}$, $\vect{b}_1 \in \Rb^{h}$ and $\vect{b}_2 \in \Rb^{nd}$.



\section{Experiments}

\begin{table*}[t!]\small
    \centering
    \setlength\tabcolsep{5pt}
    \resizebox{\textwidth}{!}{%
    \begin{tabular}{llllllllll}
     \toprule
     & \multicolumn{2}{c}{\bf{Single Sentence}} & \multicolumn{3}{c}{\bf{Sentence Similarity}} & \multicolumn{3}{c}{\bf{Natural Language Inference}} \\
     & \bf{SST-2} & \bf{COLA} & \bf{MRPC} & \bf{STS-B} & \bf{QQP} & \bf{MNLI} & \bf{QNLI} & \bf{RTE} & \bf{Avg.} \\
     \midrule
     \multicolumn{10}{l}{\it \textbf{T5-Base}}\\
Prompt Tuning & $72.37_{2.61}$ & $-8.28_{3.01}$ & $67.65/75.58_{5.17/6.12}$ & $56.98/56.56_{5.3/5.45}$ & $75.95/71.37_{0.37/0.81}$ & $58.38_{2.55}$ & $71.57_{3.47}$ & $56.28_{3.42}$ & $56.55_{1.88}$\\
SPT - LowRank & $\bf  88.94_{0.33}$ &$\bf  10.17_{14.95}$ & $\bf  86.04/89.75_{2.53/2.06}$ & $\bf  83.07/81.98_{3.76/0.68}$ & $82.32/76.6_{0.35/0.16}$ & $67.6_{0.5}$ & $81.92_{1.09}$ &$\bf  61.59_{1.25}$ &$\bf  70.01_{2.15}$\\
Fully Finetune & $86.03_{3.65}$ & $4.18_{5.57}$ & $70.77/79.72_{1.14/2.22}$ & $51.43/52.92_{6.37/5.42}$ & $\bf  88.23/84.48_{0.79/0.5}$ & $\bf  78.71_{0.48}$ & $\bf  82.06_{4.1}$ & $52.66_{1.11}$ & $64.68_{0.24}$\\
     \midrule
     \multicolumn{10}{l}{\it \textbf{T5-Base}}\\
Prompt Tuning & $63.61_{16.5}$ & $-5.11_{0.98}$ & $68.47/79.05_{2.56/3}$ & $64.91/64.99_{4.69/4.71}$ & $78.84/74.98_{1.33/1.48}$ & $70.76_{8.6}$ & $71.39_{7.03}$ & $48.79_{2.21}$ & $58.13_{3.02}$\\
SPT - LowRank & $\bf 93.26_{0.57}$ & $\bf 49.26_{3.57}$ & $\bf 87.98/90.06_{1.8/1.59}$ & $\bf86.63/86.88_{1.53/1.14}$ &$85.28/80.47_{0.54/1.05}$ & $80.94_{0.57}$ & $ 89.06_{0.23}$ &$\bf  65.94_{3.32}$ &
$\bf 79.64_{0.45}$\\
Fully Finetune & $82.4_{7.54}$ & $47.1_{3.19}$ & $83.74/88.48_{5/3.67}$ & $50.67/50.55_{14.75/14.5}$ & $\bf 90.89/87.81_{0.25/0.32}$ & $\bf 86.65_{0.21}$ & $\bf 91.3_{0.17}$ & $65.46_{5.44}$ & $74.87_{1.76}$\\
    \bottomrule
    \end{tabular}}
    \caption{\textbf{Multi-task training results on GLUE.} SPT stands for Structured Prompt Tuning. \textbf{Bold} indicates the best results. We report the mean and standard deviation over 3 runs.}
    \label{tab:multitasks}
\end{table*} 


\subsection{Datasets}
We evaluate our models and baselines, such as standard soft prompt tuning and full fine-tuning, on the GLUE~\citep{wang-etal-2018-glue} and SQuAD 1.1~\citep{squad} benchmarks. 
GLUE covers multiple tasks such as paraphrase detection (MRPC, QQP), sentiment classification (SST-2) and natural language inference (MNLI, RTE, QNLI). 
SQuAD is an extractive question answering dataset, where the answer to an input question is a span in the given context. 
Since GLUE and SQuAD do not offer publicly available test sets, for small datasets (less than 10,000 samples), we use half of the validation data as the development set and the other half as the test data; for larger datasets, we keep the original validation set as the test set and sample 1,000 examples from the training set for validation.

\subsection{Pre-trained Language Model}
Following~\citet{lester2021power}, we use T5~\citep{2020t5} for the PLM. We primarily use the adapted version of T5-v1.1\footnote{T5 v1.1 LM adapted models are initialized from T5 v1.1 and then trained for 100K additional steps using the LM objective. The checkpoints are available at \url{https://github.com/google-research/text-to-text-transfer-transformer/blob/main/released_checkpoints.md}} small and base with 60M, 220M parameters, respectively as our frozen PLM. 
While we do not apply our method to other varying sizes (Large, XL, XXL) due to the computing resource constraint, the experiments can be easily extended to other sizes. 
For a fair comparison, the full fine-tuning method is also updated from the same adapted version of T5-v1.1.

\subsection{Training Details}
We follow most of the hyperparameters and details in~\citep{lester2021power} except the optimizer, where we use AdamW~\citet{Loshchilov2019DecoupledWD} instead of Adafactor~\citep{adafactor}.
Although Adafactor reduces the memory requirements for the square gradients from square to sub-linear, it introduces new hyperparameters: decay rate, scale parameter and relative steps. 
To reduce the hyperparameter search space, we use AdamW and do not observe significant performance differences.
All of our models are reimplemented with Pytorch~\cite{pytorch} and Huggingface's Transformers library~\cite{transformers}. 
During tuning, the newly introduced parameters are the shared embedding $\vect{e}_{t}$ and hypernetwork $H$. 
In all cases, we set the prompt length $n = 20$ and task embedding dimension $k = 64$. For standard prompt tuning, we initialize the prompt tokens with the top 20 most frequent tokens in T5's vocabulary. 
For structured prompt tuning, the hypernetwork and shared embedding are drawn from the normal distribution. 

\subsection{Results}
Table~\ref{tab:main_results} presents the main results, the model performance on the GLUE and SQuAD benchmarks.
Structured Prompt Tuning (denoted as SPT) consistently outperforms the standard prompt tuning method in almost every setting, with $+1.5$/$+1.2$-point gains with T5-v1.1 small/base in average on GLUE. 
Although there are still some significant performance differences between prompt tuning and full finetuning, our structured prompt tuning methods have successfully reduced the gap.


\paragraph{Multi-Task Learning}

We also compare structured prompt tuning with other baselines in the multi-task learning setting on GLUE.
Each batch consists of roughly the same number of examples sampled from the training set of each task.
For structured prompt tuning, the hypernetwork $H$ is shared across all tasks but each task has its own unique task embedding $\vect{e}_{t}$.
This design aims for preserving the shared knowledge of multiple tasks in the hypernetwork, while allowing some flexibility for adapting to specific tasks through $\vect{e}_{t}$.
In contrast, the standard prompt tuning treats all examples the same, regardless of which task an example belongs to.
As shown in Table~\ref{tab:multitasks}, our structured prompt tuning (the low rank variant) generally performs much better than the standard prompt tuning. Perhaps more surprisingly, it also outperforms full fine-tuning on multiple tasks and has a higher average score overall.


\ignore{

Table~\ref{tab:multitasks} shows the multi-task learning results. 
In multitask training setting, the hypernetwork $H$ is shared across all tasks, and the shared embedding $\vect{e}_{t}$ is unique for each task.
It is not surprising that structured prompt tuning can outperform standard prompt tuning.
Restricted by the design of standard prompt tuning, standard prompt tuning has limited tunable parameters. When it train on multiple tasks at the same time would significantly hurt the performance. However, structured prompt tuning can be applied to multitask training. The hypernetwork allows the models share information between the tasks and achieve the comparable performance to the single-task setting.

Compared to single-task performances, the performances of both standard prompt tuning and fully fine-tune methods drop significantly. On the other hand, there is almost no difference for structured prompt tuning on T5-v1.1 base and only a small performance drop for structured prompt tuning on T5-v1.1 small. Introducing less than 100 parameters for each task, multi-task structured prompt tuning can achieve the performance almost the same as the single-task setting.

}


\ignore{

\subsection{Prompt Initialization}
\begin{table}[t!]\small
    \centering
    \setlength\tabcolsep{5pt}
    \begin{tabular}{lll}
    \toprule 
    \textbf{Model} & \textbf{SST-2} &  \textbf{STSB} \\
    \midrule
    \multicolumn{3}{l}{\it \textbf{T5-Small}}\\
    Prompt Tuning& 72.4 & 22.8\\
    Prompt Tuning-SI& 73.2 & 0\\
    SPT - Linear& 89.7 & 85.5 \\
    \midrule 
    \multicolumn{3}{l}{\it \textbf{T5-Base}}\\
    Prompt Tuning& 81.5 & 58.7\\
    Prompt Tuning-SI& 0.1 & 60.3 \\
    SPT - Linear& 92.8 & 88.5\\
    \bottomrule
    \end{tabular}
    \caption{\textbf{The results of different prompt initialization.} We run the experiments with learning rate $0.01$. With this setting, the difference between structured prompt tuning and prompt tuning with special initialization is most evident. 
    SPT stands for structured prompt tuning and SI stands for special initialization.}
    \label{tab:initialization}
\end{table} 

Is the improvement due to our assumption that hypernetworks endow non-trivial correlation to the prompt embeddings, or have we found a better way to initialize prompt embeddings? 
To answer this question, we train a standard prompt tuning model with a special initialization. 
We use a randomly initialized linear hypernetwork and task embedding to generate a set of prompt embeddings. We then use the generated prompt embeddings as initialization for prompt tuning. 
In this way, prompt tuning and structured prompt tuning generate the same cues at the beginning of training.
As we can see in Table~\ref{tab:initialization}, the results of prompt tuning with special initialization are quite different from structured prompt tuning results, but close to standard prompt tuning results. 
This shows that the hypernetwork is needed, and the improvement is not a result of initialization. A simple linear generator can yield significant gains.  

}

\paragraph{Learning Rate Sensitivity}

\begin{figure}[t!]
    \centering
    \includegraphics[width=\columnwidth]{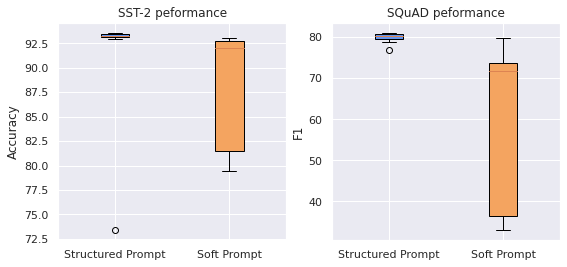}
    \caption{\textbf{The box plot of prompt tuning with different learning rates (0.01, 0.1, 1).} Structured prompt tuning (the low-rank matrix variant) is much less sensitive to learning rate than the standard prompt tuning. 
    Each learning rate has 3 runs with different random seeds.}
    \label{fig:box}
\end{figure}

\citet{li2021prefixtuning} point out that prompt tuning is very sensitive to the change of learning rate. 
However, we do not observe similar issues on structured prompt tuning.
To show this, we test three different learning rates (0.01, 0.1, 1) for both the standard and our structured prompt tuning (with the low-rank matrix hypernetwork) methods on SST-2 and SQuAD, where each learning rate has 3 runs with different random seeds.
As can be seen clearly in Figure~\ref{fig:box}, the variance of standard prompt tuning is quite large, while different learning rates have almost no effect on structured prompt tuning.
We leave more analysis on this phenomenon for future work.

\section{Conclusion}
In this work, we present \emph{structured prompt tuning}, strengthening the relationship between soft tokens via generating them with task embedding. 
Our experiments on GLUE and SQuAD show that structured prompt tuning consistently outperforms standard prompt tuning in the single-task setting. 
For multi-task settings, our proposed method outperforms both standard prompt tuning and the full fine-tuning method. 
We also found that structured prompt tuning is much less sensitive to the learning rate change.

\bibliography{custom}
\bibliographystyle{acl_natbib}



\end{document}